\documentclass[conference]{IEEEtran}
\usepackage{cite}
\usepackage{amsmath,amssymb,amsfonts}
\usepackage{algorithmic}
\usepackage{graphicx}
\usepackage{textcomp}
\usepackage{xcolor}
\usepackage[acronym]{glossaries}
\usepackage{mathtools}
\usepackage[colorlinks=false]{hyperref}
\hypersetup{
    colorlinks,
    linkcolor={black},
    citecolor={black},
    urlcolor={black}
}
\def\BibTeX{{\rm B\kern-.05em{\sc i\kern-.025em b}\kern-.08em
    T\kern-.1667em\lower.7ex\hbox{E}\kern-.125emX}}
\ifCLASSOPTIONcompsoc
\usepackage[caption=false,font=normalsize,labelfon
t=sf,textfont=sf]{subfig}
\else
\usepackage[caption=false,font=footnotesize]{subfig}
\fi
\DeclareMathOperator*{\argmin}{arg\,min}
\usepackage{diagbox}
\usepackage{gensymb}
\usepackage{url}
\begin{document}
\newacronym{ddpm}{DDPM}{Denoising Diffusion Probabilistic Model}
\newacronym{ldm}{LDM}{Latent Diffusion Model}
\newacronym{nlp}{NLP}{Natural Language Processing}
\newacronym{llm}{LLM}{Large Language Model}
\newacronym{sd}{SD}{Stable Diffusion}
\newacronym{rcs}{RCS}{Rare Class Sampling}
\newacronym{pdf}{PDF}{Probability Density Function}
\newacronym{pmf}{PMF}{Probability Mass Function}
\newacronym{kde}{KDE}{Kernel Density Estimation}
\newacronym{saln}{SALN}{Style-Adaptive Layer Normalization}
\newacronym{rbf}{RBF}{Radial Basis Function}
\newacronym{cfg}{CFG}{Classifier-free Guidance}
\newacronym{fid}{FID}{Fréchet Inception Distance}
\newacronym{sota}{SotA}{State of the Art}
\newacronym{ddim}{DDIM}{Denoising Diffusion Implicit Model}
\title{Solar Altitude Guided Scene Illumination}

\author{
\IEEEauthorblockN{Samed Do\u{g}an, Maximilian Hoh, Nico Leuze, Nicolas Rodriguez Peña, Alfred Schöttl}
\IEEEauthorblockA{Dept. of Electrical Engineering and Information Technology\\ 
University of Applied Sciences Munich, 80335 Munich, Germany\\
Email: samed.dogan@hm.edu\\}
}

\maketitle

\begin{abstract}
The development of safe and robust autonomous driving functions is heavily dependent on large-scale, high-quality sensor data. However, real-world data acquisition requires extensive human labor and is strongly limited by factors such as labeling cost, driver safety protocols and scenario coverage. Thus, multiple lines of work focus on the conditional generation of synthetic camera sensor data. We identify a significant gap in research regarding daytime variation, presumably caused by the scarcity of available labels. Consequently, we present solar altitude as global conditioning variable. It is readily computable from latitude-longitude coordinates and local time, eliminating the need for manual labeling. Our work is complemented by a tailored normalization approach, targeting the sensitivity of daylight towards small numeric changes in altitude. We demonstrate its ability to accurately capture lighting characteristics and illumination-dependent image noise in the context of diffusion models.
\end{abstract}
\section{Introduction}
Diffusion models \cite{ho20ddpm, sohl-dickstein15} have emerged as a powerful class of generative models, capable of synthesizing data samples of arbitrary modality. They consequently gained significant traction within the autonomous driving domain, enabling generation of complex driving scenarios under controlled conditions. \\
Large bodies of work focus on the realistic generation of synthetic camera \cite{gao2023magicdrive, swerdlow2024street} or multi-modal data \cite{wu2024holodrive, xdrive, wang2024drivedreamer}, addressing core problems such as coherent multi-view generation and multi-sensor alignment. However, given the extensive research effort in spatial scene control and cross-modal consistency, the aspect of accurate daytime representation remains largely unaddressed. One possible explanation is the lack of distinct labels to accurately and, in the sense of location, globally model the daytime. For instance, nuScenes \cite{nuscenes}, a dataset commonly used to train \gls{sota} generative sensor models, provides only the coarse label \textit{"Night"}. 
Another underexplored component is the limited discussion surrounding the reproducibility of the sensor hardware.  
Although research incorporates camera intrinsics and extrinsics \cite{gao2023magicdrive, wu2024holodrive}, temporal variations in sensor noise from ambient light changes remain unaccounted for, reinforcing the relevance of this work. \\
We present a methodology that tackles previously outlined challenges and can be incorporated into existing setups in an orthogonal manner. First, we introduce solar altitude as a novel global light conditioning variable. To the best of our knowledge, we are the first to do so within the domain of autonomous driving. Solar position can be directly derived from the dataset, provided that ego-pose and capture time are available, eliminating the need for external coarse labels. We discuss the non-linear behavior between solar movement and perceived light, particularly during sunset and sunrise, and suggest a normalization approach that retains global structure and local variation within the scalar range. 
We further demonstrate the effectiveness of solar altitude as a surrogate variable in modeling perceived image noise.
Due to its lightweight hardware requirements, we use textual inversion \cite{textual-inversion} as our preferred implementation method. \\
In short, our contributions are as follows:
\begin{itemize}
    \item We introduce the task of daylight variation for generative camera sensor models in the context of autonomous driving. 
    \item We present solar altitude as a conditioning variable for global illumination and suggest a tailored normalization and encoding procedure that accounts for the non-linear relationship between conditioning and apparent brightness.
    \item We showcase the capability of altitude conditioned models to capture both global-lighting and illumination-dependent noise characteristics.
\end{itemize}
\section{Related Work}
\subsection{Diffusion Models}
Diffusion models, designed to learn some data distribution $q(\mathbf{x})$, fundamentally operate under a two-step probabilistic framework. The forward process gradually corrupts some input data $\mathbf{x}_0 \sim q(\mathbf{x}_0)$ with Gaussian noise over a series of $T$ time steps, such that $\mathbf{x}_T \sim \mathcal{N}(\mathbf{0}, \mathbf{I})$. The reverse process consists of a learnable model that recovers $\mathbf{x}_0$ from $\mathbf{x}_T$ by iterative denoising. In its canonical form, \glspl{ddpm} \cite{ho20ddpm} define the forward process as a fixed discrete-time Markov chain of length $T$, and the reverse process as a series of denoising autoencoders $\boldsymbol{\epsilon}_\theta(\mathbf{x}_t, t)$, predicting the added noise $\boldsymbol{\epsilon}$ at time step $t\sim\mathcal{U}(\{1,\dots,T\})$. \glspl{ldm} \cite{rombach2022ldm}, a class of \glspl{ddpm}, perform diffusion in the latent space, given some latent variable $\mathbf{z}=\mathcal{E}(\mathbf{x})$, where $\mathcal{E, D}$ is a pre-trained encoder-decoder pair, respectively. Including some conditioning $\mathbf{c}$ of arbitrary modality, $\boldsymbol{\epsilon}\sim\mathcal{N}(\mathbf{0}, \mathbf{I})$ and $\mathbf{z}_0=\mathcal{E}(\mathbf{x}_0)$, the \gls{ldm} training objective is defined as:
\begin{equation}
    \label{eqn: ldm_loss}
    L_{\gls{ldm}} \coloneqq \mathbb{E}_{\mathbf{z}_0, \boldsymbol{\epsilon}, t, \mathbf{c}}\left[\|\boldsymbol{\epsilon}-\boldsymbol{\epsilon}_\theta(\mathbf{z}_t, t, \mathbf{c})\|^2_2\right] . 
\end{equation}
Equation \eqref{eqn: ldm_loss} is a simplified version of the variational lower bound discussed in \cite{ho20ddpm}.
\subsection{Textual Inversion}
\begin{figure}[!t]
    \centering
    \includegraphics[width=\linewidth]{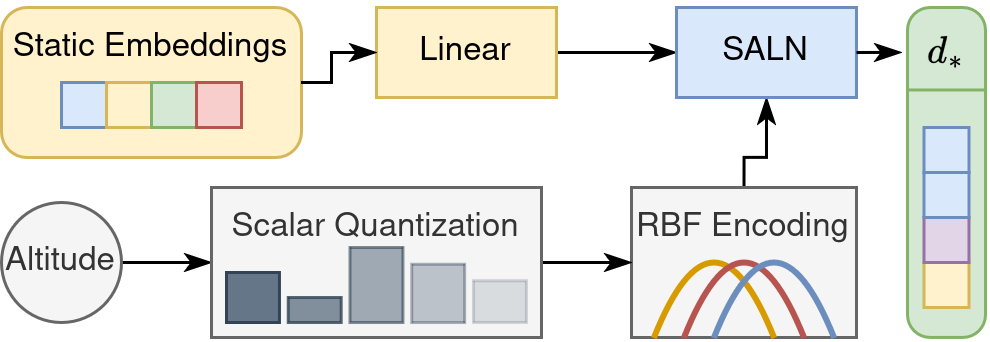}
    \caption{Dynamic Embedding Pipeline}
    \label{fig:network_architecture}
\end{figure}
\begin{figure*}[t]
\centering
\subfloat{\includegraphics[width=0.23\linewidth]{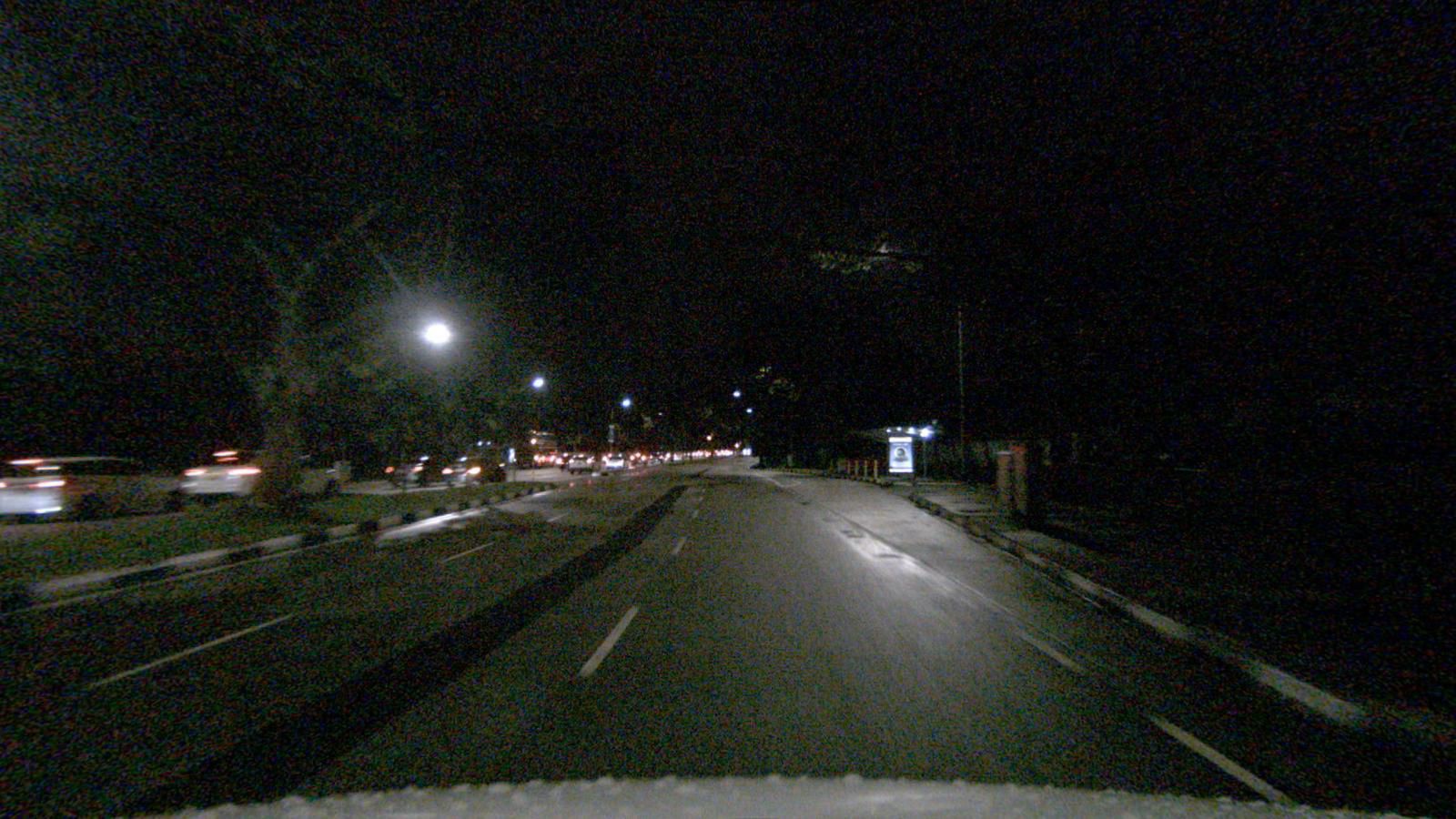}}
\hfil
\subfloat{\includegraphics[width=0.23\linewidth]{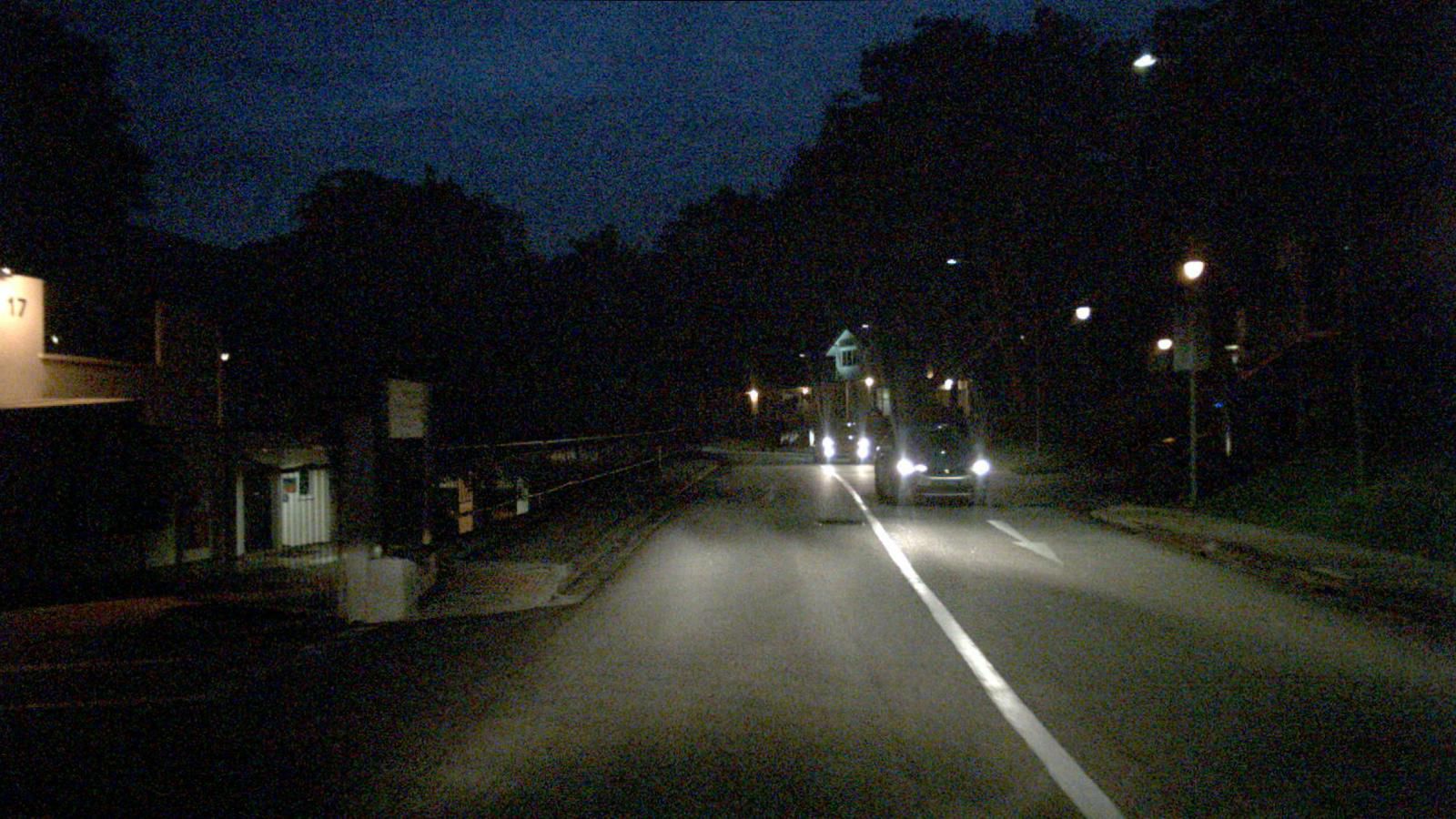}}
\hfil
\subfloat{\includegraphics[width=0.23\linewidth]{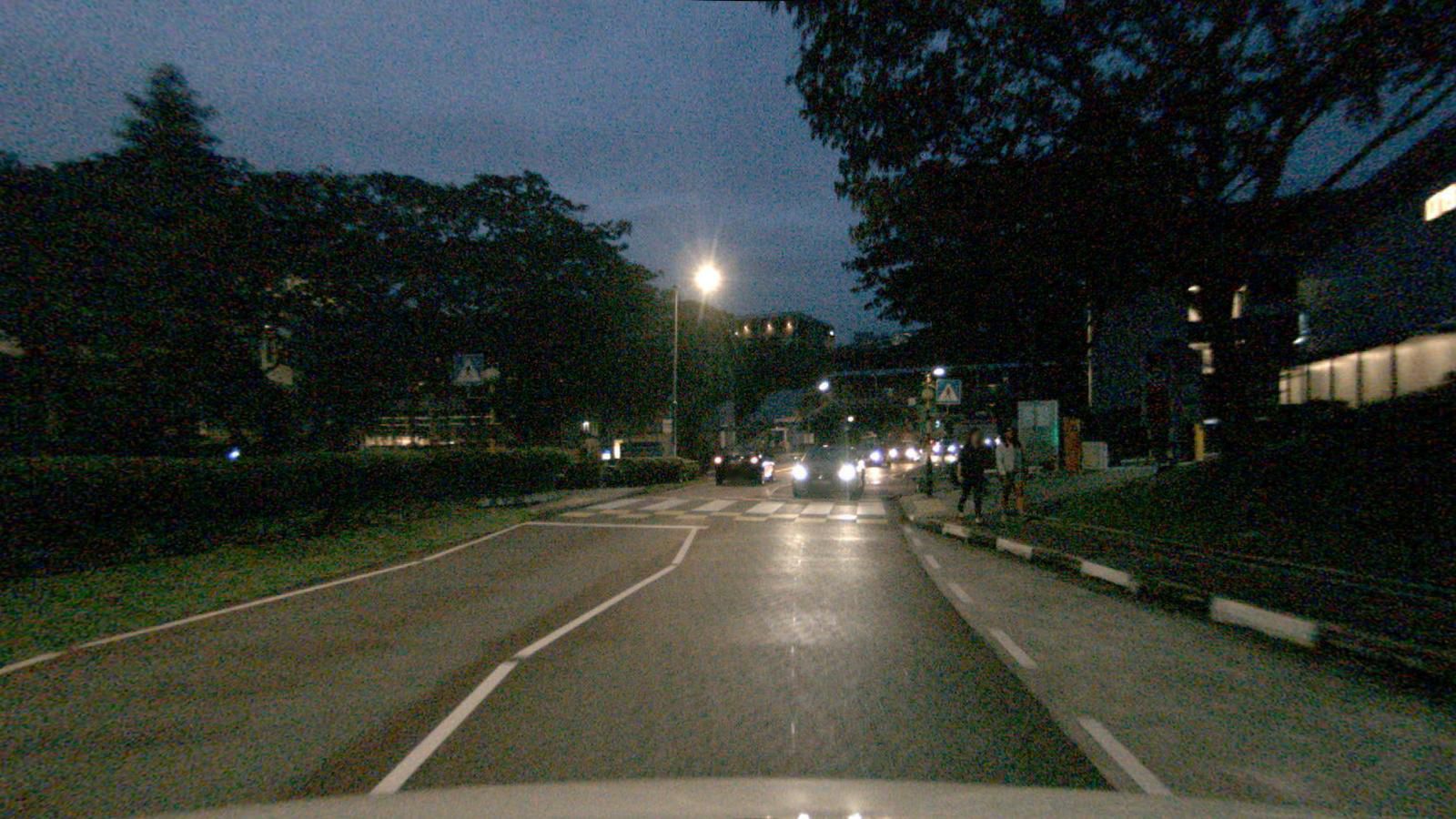}}
\hfil
\subfloat{\includegraphics[width=0.23\linewidth]{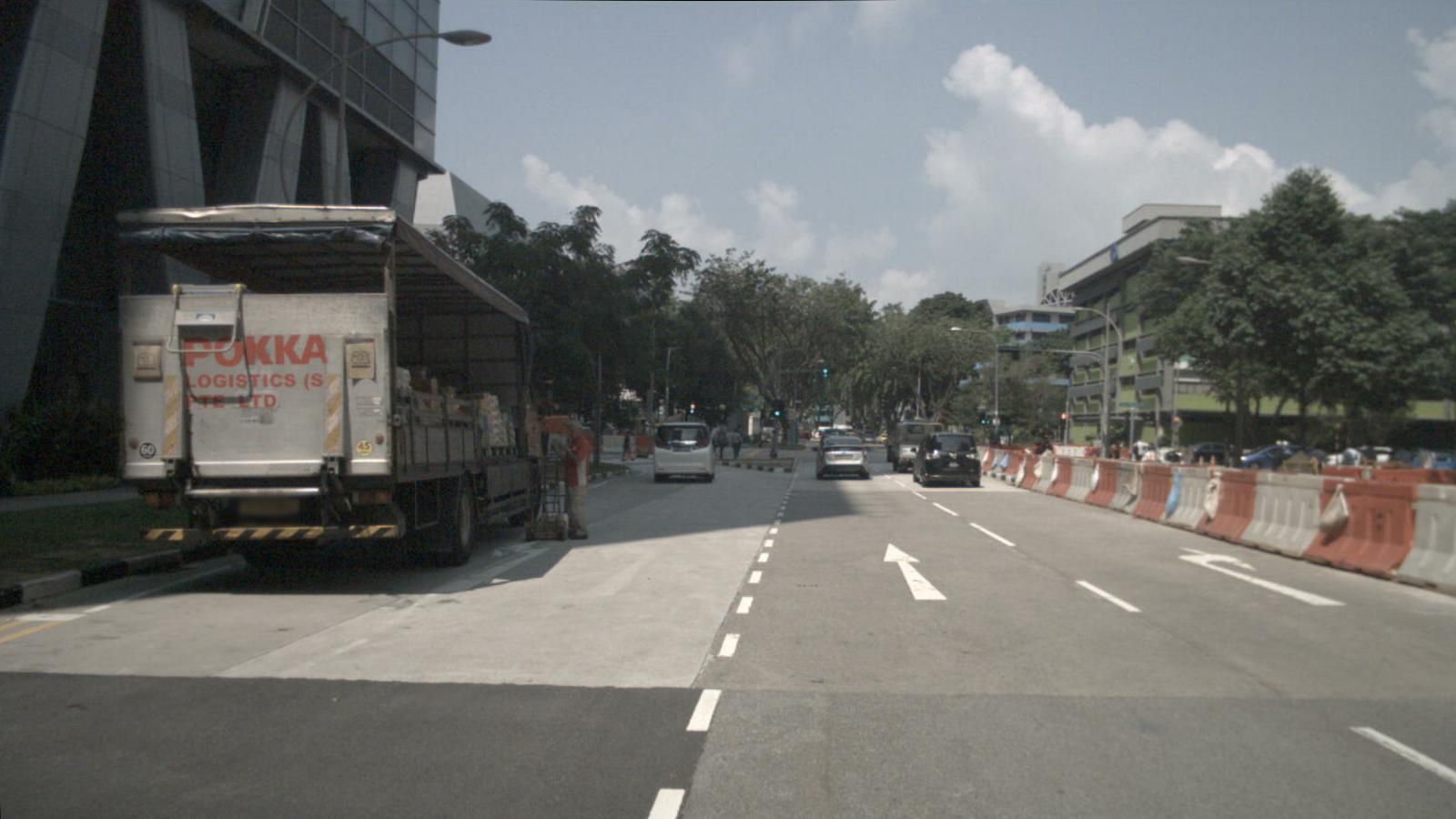}}
\vspace{-6pt}
\subfloat[0.0]{\includegraphics[width=0.23\linewidth]{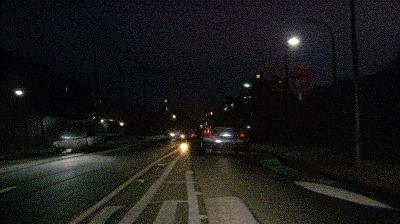}}
\hfil
\subfloat[0.33]{\includegraphics[width=0.23\linewidth]{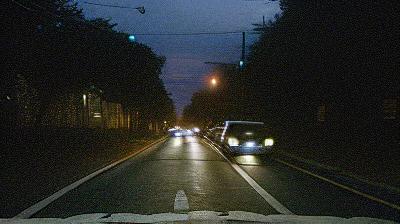}}
\hfil
\subfloat[0.66]{\includegraphics[width=0.23\linewidth]{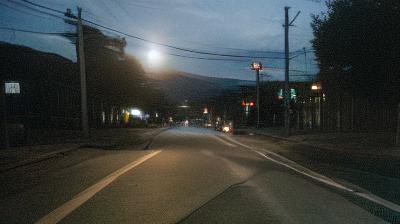}}
\hfil
\subfloat[1.0]{\includegraphics[width=0.23\linewidth]{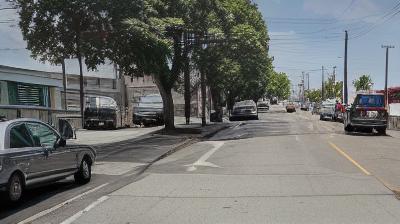}}
\caption{Comparison of ground truth samples (top) with generated images (bottom) for varying altitude conditions, normalized across the training bin range.}
\label{fig: examples}
\end{figure*}
In general, text-to-image models employ a pre-trained text encoder in order to guide the image generation process towards some desired text prompt. A tokenizer maps the input prompt to unique tokens on a per-word or sub-word basis. The discrete tokens index a corresponding learned embedding from an internal lookup table, forming the input embeddings. These are then further contextualized by the text encoder to capture sentence-level meaning.
Textual inversion \cite{textual-inversion} is a prompt-based embedding technique that injects a new concept into the vocabulary of the text encoder by introducing a \textit{pseudo-word} \textit{"V*"} \cite{pseudo-word} that corresponds to a new learnable embedding $v_*$. The pseudo-word, though devoid of inherent linguistic meaning, can be used like an ordinary word, allowing the model to integrate the new concept seamlessly into natural language. The new embedding is trained to represent the visual concept within the language domain of the diffusion model. 
\section{Methodology}
\subsection{Solar Altitude as Lighting Prior}
\noindent \textbf{Dataset Preparation.} 
Assessing camera noise and natural lighting in non-controlled environments presents several key challenges: First, diffuse lighting conditions vary greatly over time and space. Second, we need a scene-invariant surrogate variable that accurately captures the relationship between illumination and noise. Assuming constant weather, solar altitude is a valuable proxy for the expected lighting characteristics and can thus serve as a strong surrogate measure of expected image noise. Following \cite{gao2023magicdrive, xdrive, wu2024holodrive, wang2024drivedreamer}, we use nuScenes as exemplary dataset and filter samples labeled with \textit{"Rain"}, as droplets on camera introduce heavy distortion.
We convert the ego-pose to location coordinates and compute solar altitude relative to the observer horizon using the astropy library \cite{astropy}.\\

\noindent \textbf{Scalar Quantization with Residual Encoding.}
We observed a non-linear relationship between perceived light and solar position. Specifically, changes in solar altitude during sunset and sunrise cause rapid shifts in illumination, while values near the zenith have negligible influence and are broadly overshadowed by weather and external conditions. Regular normalization falls short of capturing fine variation, resulting in heavy semantic compression around sparse key regions. We draw inspiration from popular encoding schemes like positional \cite{vaswani2017attention} and Fourier encoding \cite{fourier}, aiming to preserve both coarse regions and fine grained variation within the normalization range.
Formally, given an increasing series of bin edges $b_0 < b_1 <,\dots b_{K}$ that define a range of intervals $[b_0, b_1), [b_1, b_2), \dots,[b_{K-1},b_{K})$, we first quantize the altitude $a\in \mathbb{R}$ such that $i = Q(a) \coloneqq \max\{i \in \{0, \dots, K-1\}| b_i \le a\}$. The residual function $R(a) \in [0, 1)$ captures the intra-bin variation, defined as the normalized offset of $a$ within the bin $i$: $R(a)\coloneqq\frac{a -b_i}{b_{i+1}-b_i}$. The normalized scalar $\hat{a}$ is then obtained by:
\begin{equation}
    \label{eq: norm}
    \hat{a} = N(a)\coloneqq\frac{Q(a)+R(a)}{K}.
\end{equation}
We use aggressive binning to address the limited availability of twilight and lack of sunrise samples. We further employ resampling with replacement, such that the bin count across all bins is approximately uniform. The explicit binning scheme is detailed in subsection \ref{chp:implementation_details}. 
\subsection{Token Generation}
Our method follows a two-stage process: first, we train a token that captures the general domain and visual composition of the camera images, hereafter referred to as structure token. Second, a separately trained conditional token captures the variation in daylight. We employ partial sampling similar to \cite{mou2024t2i, multistep_inversion} and use the context token during the first and the structure token during the second half of sampling. \\

\noindent \textbf{Structure Token.}
Following \cite{textual-inversion}, we learn a domain-specific word embedding $s_*$ corresponding to the pseudo-word \textit{"S*"}. Let $\mathbf{c}_\theta$ be a pre-trained CLIP \cite{clip_encoder} text encoder and let $y$ be a text prompt. We omit the neural context text-templates from \cite{textual-inversion} and \cite{clip_encoder} during training, as we observed a degradation and overarching stylization of the camera data when used in unison. As a result, our training prompt $y$ consists solely of \textit{"S*"}. The embedding $s_*$ is obtained by freezing $\boldsymbol{\epsilon}_\theta$ and $\mathbf{c}_\theta$ and directly optimizing the reconstruction objective of \eqref{eqn: ldm_loss}:
\begin{equation}
    \label{eq: textual_inversion}
    s_* = \argmin_s \mathbb{E}_{\mathbf{z}_0, \boldsymbol{\epsilon}, t, y}\left[\|\boldsymbol{\epsilon}-\boldsymbol{\epsilon}_\theta(\mathbf{z}_t, t, \mathbf{c}_\theta(y))\|^2_2\right] .
\end{equation}
We train $s_*$ on the mini subset and explicitly exclude scenes at night to mitigate the impact of poor image quality on embedding performance.\\

\noindent \textbf{Context Token.}
Contrary to the structure token, the embeddings $d_*$ corresponding to \textit{"D*"} are generated dynamically given the altitude $a$ and a small network $f_\theta$. 
To facilitate smoothness and localized sensitivity within the feature space, the scalar condition is first normalized according to \eqref{eq: norm} and enriched using an \gls{rbf} encoding layer \cite{rbf} with non-trainable uniformly placed centers and trainable gammas. We then use \gls{saln} \cite{saln} to modulate a set of static embeddings based on the enriched input condition. A complete description of our network architecture  is given in \figurename\nobreakspace\ref{fig:network_architecture}. The number of context embeddings are matched to that of the best structure embedding.
Adopting the notation from \eqref{eq: textual_inversion}, we train $f_\theta$ on the full dataset:
\begin{equation}
    \label{eqn: sunposloss}
    L_f \coloneqq \mathbb{E}_{\mathbf{z}_0, \boldsymbol{\epsilon},t,y,a}\left[\|\boldsymbol{\epsilon}-\boldsymbol{\epsilon}_\theta(\mathbf{z}_t, t, \mathbf{c}_\theta(y, f_\theta(a)))\|^2_2\right] .
\end{equation}
\newpage
\begin{table}[t]
    \centering
    \caption{Quantitative results of generated images with varying solar altitudes. Normalization is identical to \figurename\nobreakspace\ref{fig: examples}. Increments $\Delta\sigma$ are computed from left column to right column.}
    \label{tab:results}
    \begin{tabular}{l|c|c|c|c}
    \hline
    \backslashbox{Metric}{Altitude} & 0.0 &  0.33 & 0.66 & 1.0\\
    \hline\hline
    &&&&\\
    \glsentryshort{fid} (S*) \(\downarrow\) & ---& --- & --- & 151.433\\
    \glsentryshort{fid} (D* \& S*) \(\downarrow\)& 145.312 & 122.933 & 135.778 & 144.789 \\
    $\Delta\sigma$ &---& -0.137 & 0.505 & -1.223\\
    $\Delta\sigma$(GT) &---& -0.191 & 0.365 & -1.389
    \end{tabular}  
\end{table}
\section{Results}
\subsection{Implementation Details}\label{chp:implementation_details}
We implement our method over \gls{sd} $1.5$ using the diffusers library \cite{von_platen_diffusers_2022}. 
All runs are trained for 5 epochs with a batch size of $8$ and a resolution of $224\times400$, using AdamW \cite{loshchilov2017decoupled} with a weight decay of $0.01$ and learning rates of $\{0.001, 0.005, 0.0001\}$. The number of embeddings ranges from $1$ to $5$. For evaluation, we sample $64$ images per prompt using the \gls{ddim} \cite{ddim} scheduler with $30$ steps, a partial sampling step of 15 and \gls{cfg} value of $7.5$. For normalization, we employ $[a_{\text{min}}, -6, -4, -2, a_{\text{max}}]$ as bin intervals, where $a_{\text{min}}$ and $a_{\text{max}}$ represent the smallest and largest available altitudes, respectively. If applicable, we recommend a $2\degree$ step scheme from $-12\degree$ (nautical twilight) to at least $+6\degree$ (photographic golden hour).
\subsection{Evaluation}
\noindent \textbf{Structure Token.}
Similar to \cite{textual-inversion}, we use a dual score approach to assess the quality of latent space embeddings across different learning rates and token counts. In terms of visual fidelity, we compute the \gls{fid} between the training set and images generated using the prompt "\textit{S*}". Regarding textual adherence, we measure the average cosine similarity between image and text embeddings in BLIP\cite{li2022blip} space. Specifically, we generate images with "\textit{S*} at night." and compare them with a variety of image captions semantically close to "A traffic scene at night". We select the best performing embedding for inference.\\

\noindent\textbf{Context Token.} 
The validation set does not contain samples captured during twilight conditions, rendering direct comparison infeasible. Instead, we evaluate the context token based on its ability to faithfully generate images across the provided altitude. We additionally measure the rate of change in estimated noise $\Delta\sigma$ using \cite{noise2}. Ground truth data is filtered for back and front views. We complement the quantitative results in Table \ref{tab:results} with qualitative examples provided in \figurename\nobreakspace\ref{fig: examples}.
\section{Conclusion}
This work highlights current limitations in fine-grained control over scene illumination in the context of camera data generation. We hypothesize that limited research is due to the constrained availability of relevant labels. To this end, we introduce solar altitude as a control variable that can be seamlessly integrated into existing diffusion frameworks. The computation is fully automated, requiring no manual labels, underscoring its scalability for large-scale industrial datasets. We suggest a tailored normalization approach that emphasizes numerical variation in perceptually sensitive regions.
Despite the very limited capacity of textual inversion, compounded by the scarcity of relevant data, our approach shows promising results. Future work may incorporate solar azimuth alongside altitude, enabling simulation of driver and sensor glare, accurate shadow rendering, and improved robustness of driving functions by evaluating identical scenes under systemically varied illumination.
We are confident that integration into large-scale models will further improve precision and generalization. As such, we believe this work provides an additional stepping stone towards generating realistic and controllable camera data.
\section*{Acknowledgments}
The research leading to these results is funded by the German Federal Ministry for Economic Affairs and Energy within the project “NXT GEN AI METHODS – Generative Methoden für Perzeption, Prädiktion und Planung" (grant no. 19A23014M).
\newpage
\bibliographystyle{IEEEtran}
\bibliography{references}
\end{document}